\documentclass{article}

\usepackage[margin=0.75in]{geometry}   
\usepackage{times}                     
\usepackage{microtype}                 

\usepackage{indentfirst}

\usepackage{amsmath,amssymb}

\usepackage{graphicx}
\usepackage{subcaption}
\usepackage{booktabs}
\usepackage{multirow}
\usepackage{multicol}
\usepackage{float}
\usepackage{stfloats}
\usepackage{placeins}
\usepackage{afterpage}

\usepackage{algorithm}
\usepackage{algpseudocode}

\usepackage{caption}
\DeclareCaptionFont{algsmall}{\fontsize{8}{9}\selectfont}
\usepackage[numbers]{natbib}
\usepackage{hyperref}

\usepackage{sectsty}
\sectionfont{\large\bfseries}
\subsectionfont{\normalsize\bfseries}
\subsubsectionfont{\normalsize\itshape}

\usepackage[margin=0.75in]{geometry}
\usepackage{authblk}
\title{\bfseries \Large Prior Distribution and Model Confidence}

\author[1]{Maksim Kazanskii}
\author[2]{Artem Kasianov}

\affil[1]{Independent Researcher, \texttt{mkazanskii@gmail.com}}
\affil[2]{BIOPOLIS/CIBIO, Vairão, Portugal}

\date{}
\begin{document}

\twocolumn[
\begin{@twocolumnfalse}
 \maketitle

  \vspace{1em}
\end{@twocolumnfalse}
]

\begin{center}
\section*{Abstract}
\end{center}

We study how the training data distribution affects confidence and performance in image classification models. We introduce Embedding Density, a model-agnostic framework that estimates prediction confidence by measuring the distance of test samples from the training distribution in embedding space, without requiring retraining. By filtering low-density (low-confidence) predictions, our method significantly improves classification accuracy. We evaluate Embedding Density across multiple architectures and compare it with state-of-the-art out-of-distribution (OOD) detection methods. The proposed approach is potentially generalizable beyond computer vision.
\newline

\section{Introduction}
 
In recent years, deep learning has driven progress in computer vision, enabling robust performance across tasks such as image classification, object detection, and semantic segmentation \cite{krizhevsky2012imagenet,he2016resnet,long2015fcn}. Despite these advances, modern visual recognition systems remain highly sensitive to violations of the closed-set assumption, where test inputs deviate from the distribution of the training data due to changes in environment, acquisition conditions, or the presence of previously unseen semantic categories \cite{scheirer2013openset,bendale2015opensetdeep,yang2024good}. Under such shifts of distribution, models often produce confident but incorrect predictions, raising critical concerns for the reliability of vision-based systems in real-world and safety-critical applications \cite{amodei2016ai,hendrycks2019robustness,sayyed2025ood}.

From a theoretical perspective, the behavior of deep visual models under distribution shift is tied to the geometry and structure of the latent representations induced by the training data. The distribution of features in this representation space defines the effective domain of the model and constrains the regions in which its predictions can be considered meaningful \cite{bengio2013representation,arora2019contrastive,kirichenko2023lastlayer,zhou2024dgsurvey,lu2025taskood}. Understanding and using this structure offers a path toward characterizing model predictability beyond empirical confidence measures.

In this work, we propose an approach for out-of-distribution (OOD) detection in computer vision that leverages the latent space geometry of training data to assess the reliability of model predictions at the instance level. Our method operates without additional supervision or retraining and is applicable across a range of visual tasks and network architectures. We evaluate the proposed framework on standard large-scale OOD benchmarks and under realistic distribution shifts, demonstrating consistent improvements over existing methods \cite{hendrycks2022scalingood}.

\section{Related Work}
Out-of-distribution (OOD) detection in computer vision has been studied in the context of deep neural networks for visual recognition. Early approaches rely on confidence-based criteria, such as thresholding the maximum softmax probability, and input perturbation and temperature scaling as in ODIN, which improve separability between in-distribution and OOD samples but often require dataset-specific calibration \cite{hendrycks2017baseline, liang2018odin, bitterwolf2023inorout, jelenic2024smoothness}.
Feature-space methods model the distribution of learned representations, for example through class-conditional Gaussian assumptions and Mahalanobis distance, enabling a unified treatment of OOD detection and adversarial examples while remaining sensitive to feature dimensionality and covariance estimation \cite{lee2018mahalanobis, bitterwolf2023inorout, jelenic2024smoothness}.
Energy-based approaches reinterpret model outputs as unnormalized log-densities, providing a scoring function that generalizes across architectures and large-scale datasets \cite{liu2020energy, yang2024survey}.

Uncertainty estimation methods, such as deep ensembles, approximate predictive uncertainty by aggregating predictions from multiple independently trained models, yielding strong empirical performance at the cost of increased computational and memory overhead \cite{lakshminarayanan2017ensembles, ovadia2019trust, fang2023ensemble}.
More recent work explores contrastive and self-supervised representation learning to induce feature spaces that improve robustness under distribution shift \cite{tack2020csi, aathreya2024flowcon}.
In contrast, our approach focuses on the intrinsic geometry of the latent representation induced by the training distribution, enabling instance-level reliability assessment without auxiliary outlier data, architectural modification, or ensemble training.

\section{Methods}

To study the effect of data distribution on model confidence, we selected several models trained on the same dataset. All models were trained on the ImageNet-1K dataset \citep{deng2009imagenet}, ensuring a consistent training data distribution across experiments. To analyze differences in our framework's performance, we intentionally included embedding architectures from different model families with varying parameter counts. Specifically, we considered convolutional ResNet models (ResNet-50 and ResNet-101 \citep{he2016resnet}), Vision Transformer–based models \citep{touvron2021deit} (DeiT-T, DeiT-S, and DeiT-B), and the lightweight ShuffleNet-V2 model \citep{ma2018shufflenetv2}. A brief summary of the selected models is provided in Table~\ref{tab:models}.

\begin{table}[h]
\centering
\caption{Overview of the models used in our experiments. All models were trained on the ImageNet-1K dataset to ensure consistent data distribution.}
\resizebox{\columnwidth}{!}{%
\begin{tabular}{l l l l r}
\toprule
\textbf{Model} & \textbf{Training Dataset} & \textbf{Architecture Type} & \textbf{Reference} & \textbf{\# Parameters} \\
\midrule
ResNet-101      & ImageNet-1K & Convolutional (CNN) & \citep{he2016resnet}        & 44.5M \\
ResNet-50       & ImageNet-1K & Convolutional (CNN) & \citep{he2016resnet}        & 25.6M \\
ShuffleNet-V2   & ImageNet-1K & Lightweight CNN     & \citep{ma2018shufflenetv2}  & 2.3M  \\
DeiT-Tiny       & ImageNet-1K & Vision Transformer (ViT) & \citep{touvron2021deit} & 5.7M  \\
DeiT-Small      & ImageNet-1K & Vision Transformer (ViT) & \citep{touvron2021deit} & 22.1M \\
DeiT-Base       & ImageNet-1K & Vision Transformer (ViT) & \citep{touvron2021deit} & 86.4M \\
\bottomrule
\end{tabular}%
}

\label{tab:models}
\end{table}

To characterize the data distribution on which the models were trained, we analyzed the structure of the ImageNet-1K training set by computing image embeddings using several pretrained models. We selected a diverse set of architectures to generate embedding representations for each image in the dataset, enabling an examination of the statistical properties of the data across different feature spaces. This embedding-based analysis provides insight into how different models encode the visual information present in the training distribution.

We use only the training split of ImageNet-1K, as the models listed in Table~\ref{tab:models} were trained exclusively on this subset. For the embedding analysis, we employed the following models: DINO-v1 \citep{caron2021emerging}, DINO-v2 \citep{oquab2023dinov2} with embedding dimensions of 384, 768, and 1024, and MobileNet-V2 \citep{sandler2018mobilenetv2}.

The selection of embedding models was guided by two main factors: model capacity and the nature of the data used for pretraining. To evaluate the sensitivity of our algorithm to embedding dimensionality, we compared performance across closely related models with different embedding sizes. Additionally, we investigated how differences in pretraining data influence the resulting embeddings. A brief summary of the embedding models is provided in Table~\ref{tab:embedding_models}.
\begin{table}[h]
\centering
\caption{Description of the models and their embedding sizes.}
\resizebox{\columnwidth}{!}{%
\begin{tabular}{l l l l}
\toprule
\textbf{Model} & \textbf{Training Data} & \textbf{Embedding Size} & \textbf{Reference} \\
\midrule
DINO-V1 ViT-S/16   & ImageNet-1K (self-supervised)         & 384   & [4] \\
DINO-V1 ViT-B/16   & ImageNet-1K (self-supervised)         & 768   & [4] \\
DINO-V1 ViT-B/8    & ImageNet-1K (self-supervised)         & 768   & [4] \\
DINO-V2 ViT-S/14   & 142M curated dataset (no labels)      & 384   & [5] \\
DINO-V2 ViT-B/14   & 142M curated dataset (no labels)      & 768   & [5] \\
MobileNet-V2       & ImageNet-1K (supervised)              & 1000  & [6] \\
\bottomrule
\end{tabular}%
}

\label{tab:embedding_models}
\end{table}

In order to investigate the effect of the prior data distribution on the prediction capabilities of the classification models, we selected two distinct datasets:
ImageNet-V2 \citep{recht2019imagenetv2} and ObjectNet \cite{barbu2019objectnet}.  We divided ImageNet-V2 into training and internal test sets with a split of 75\% / 25\% and left the ObjectNet dataset as the external test set. We used only ObjectNet classes that coincide with the classes in the original ImageNet-1K dataset (yielding approximately $10k$ images). We used the training dataset to tune two parameters: the number of neighbors $N$ and the distance threshold $L$.  We divided the test sets (external and internal) into three label-disjoint subsets. Therefore, images with the same label can only be present in one subset. The goal of this division is to report variability in the metrics.

To compute embedding similarities, we used cosine similarity \citep{manning2008introduction}, a commonly adopted metric for nearest-neighbor search in high-dimensional spaces. We further assessed the robustness of our results to alternative distance metrics. Embeddings were stored and queried using ChromaDB \citep{chromadb} with a ClickHouse backend \citep{clickhouse}, enabling efficient large-scale retrieval. All experiments were performed on a MacBook equipped with an Apple M1 processor, using either the MPS or CPU backend.

Confidence estimation was performed on both the training and test sets. On the training set, the parameters 
$N$ and 
$L$ were tuned to optimize performance. The optimized parameters were then applied to the test sets (internal and external), on which performance was reported. The pseudocode for the algorithm is presented below.

\begingroup
\scriptsize
\begin{algorithm}[H]
\fontsize{8}{10}\selectfont
\caption{Confidence Estimation via Embedding Density Filtering (Single Embedding Model)}
\begin{algorithmic}[1]

\Require Base set ($N_{\text{base}}$), training set ($N_{\text{train}}$), test set ($N_{\text{test}}$),
embedding model ($m$), classifier ($C$), confidence threshold ($L$),
neighbor count threshold ($N$), normalized cumulative gain ($\mathrm{NCG}$),
and distance function $d(\cdot,\cdot)$.

\vspace{0.5em}
\Statex \textit{Build and store embedding index $E$ (images from the Base set)}
\vspace{0.5em}

\ForAll{$x \in N_{\text{base}}$}
    \State $E(x) \gets m(x)$
    \State $E \gets E \cup \{E(x)\}$
\EndFor

\vspace{0.5em}
\Statex \textit{Compute neighbors for tuning and tune density thresholds}
\vspace{0.5em}

\ForAll{$x \in N_{\text{train}}$}
    \State $\hat{y} \gets C(x)$
    \State $\mathcal{C}_x \gets \{(N_i(x), L_i(x))\}_{i=1}^{K_{\max}}$
    \Statex \Comment{Top-$K_{\max}$ neighbors of $m(x)$ in $E$}
    \State Record $(c_x,\; \mathbf{1}[\hat{y}=y])$
\EndFor

\vspace{0.5em}
\Statex \textit{Find optimal parameters}
\vspace{0.5em}

\State $N^*, L^* \gets \arg\max_{N, L} \mathrm{NCG}(N, L)$

\vspace{0.5em}
\Statex \textit{Evaluate on $N_{\text{test}}$}
\vspace{0.5em}

\ForAll{$x \in N_{\text{test}}$}
    \State $\text{conf} \gets \textbf{false}$
    \State $\mathcal{N}_x \gets \text{top-}N^*\text{ neighbors of } m(x)$
    \State $c \gets \left|\left\{ z \in \mathcal{N}_x :
        d\!\left(z, m(x)\right) \le L^* \right\}\right|$
    \If{$c \ge N^*$}
        \State $\text{conf} \gets \textbf{true}$
    \EndIf
    \State $\hat{y} \gets C(x)$
    \State $\text{Res} \gets \text{Res} \cup \{(x, \hat{y}, \text{conf})\}$
\EndFor

\State \Return Res

\end{algorithmic}
\end{algorithm}
\endgroup
\begin{figure*}[t]

  \centering
 
  \begin{subfigure}{0.49\textwidth}
    \centering
    \includegraphics[width=\linewidth]{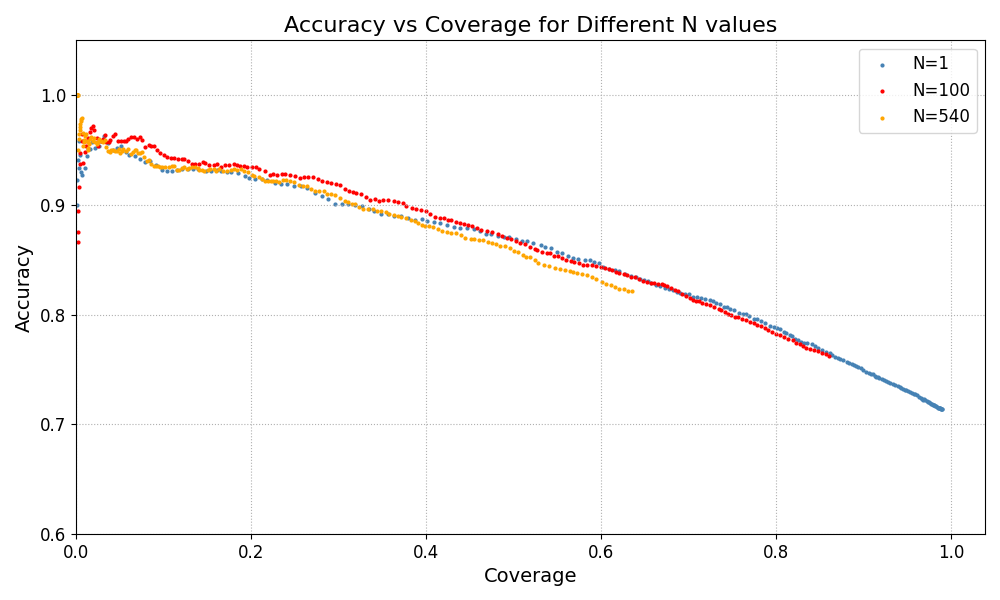}
    \caption{The Confidence Curve}
    \label{fig:fig_left}
  \end{subfigure}
  \hfill
  \begin{subfigure}{0.49\textwidth}
    \centering
    \includegraphics[width=\linewidth]{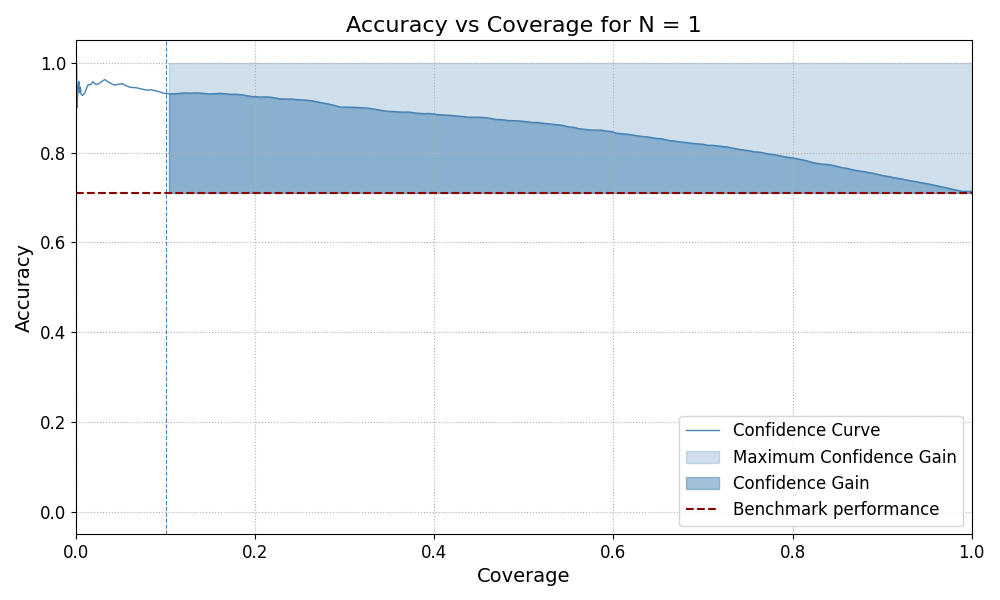}
    \caption{The Confidence Curve}
    \label{fig:fig_right}
  \end{subfigure}

   \caption{The Confidence Curves and Confidence Gain for the ResNet-101 \&  DINO-V2 ViT-B/14.}
  \label{fig:base_plot}
\end{figure*}
\afterpage{%
\begin{figure*}[b]

  \centering
 
  \begin{subfigure}{0.49\textwidth}
    \centering
    \includegraphics[width=\linewidth]{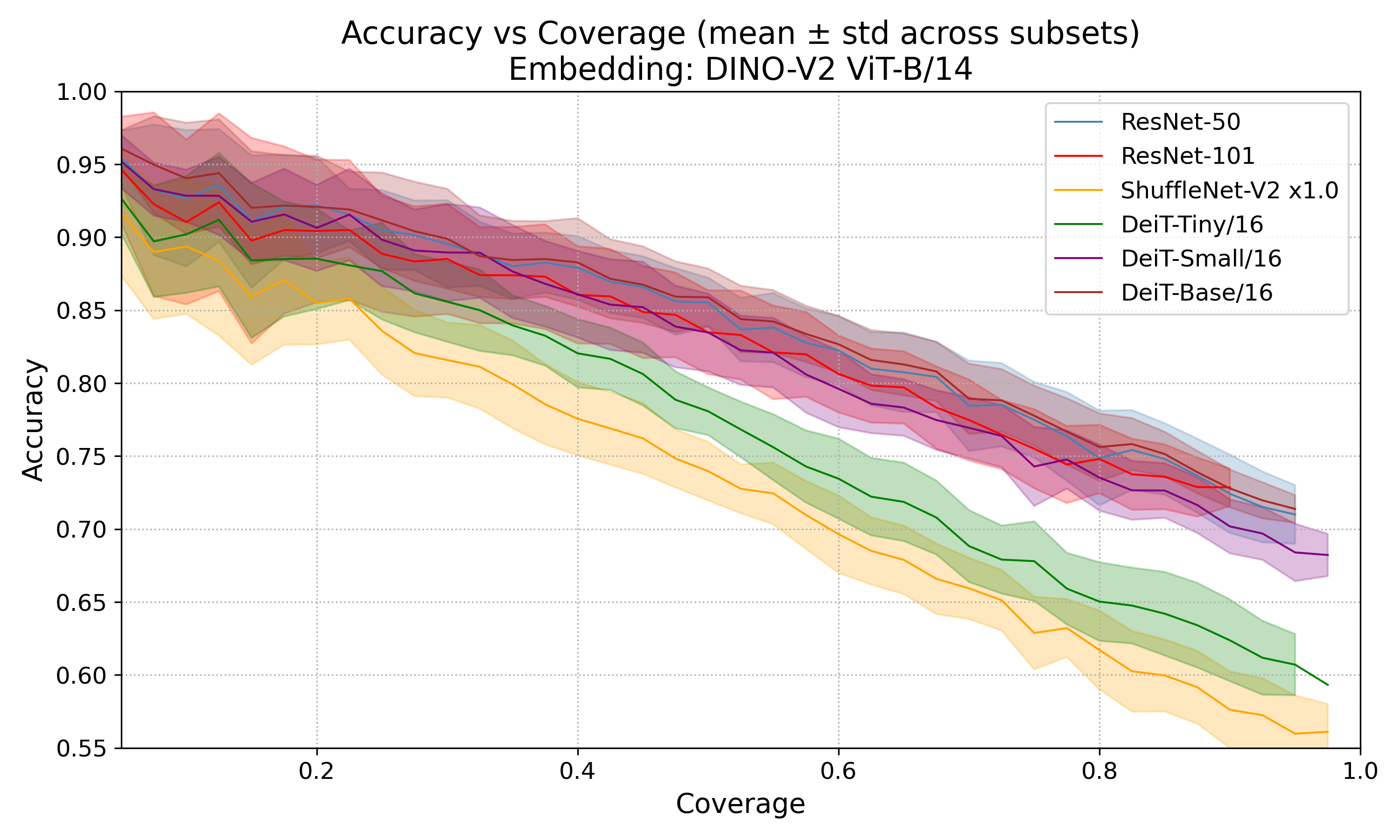}
    \caption{Confidence Curves for Internal Dataset.}
    \label{fig:fig_left}
  \end{subfigure}
  \hfill
  \begin{subfigure}{0.49\textwidth}
    \centering
    \includegraphics[width=\linewidth]{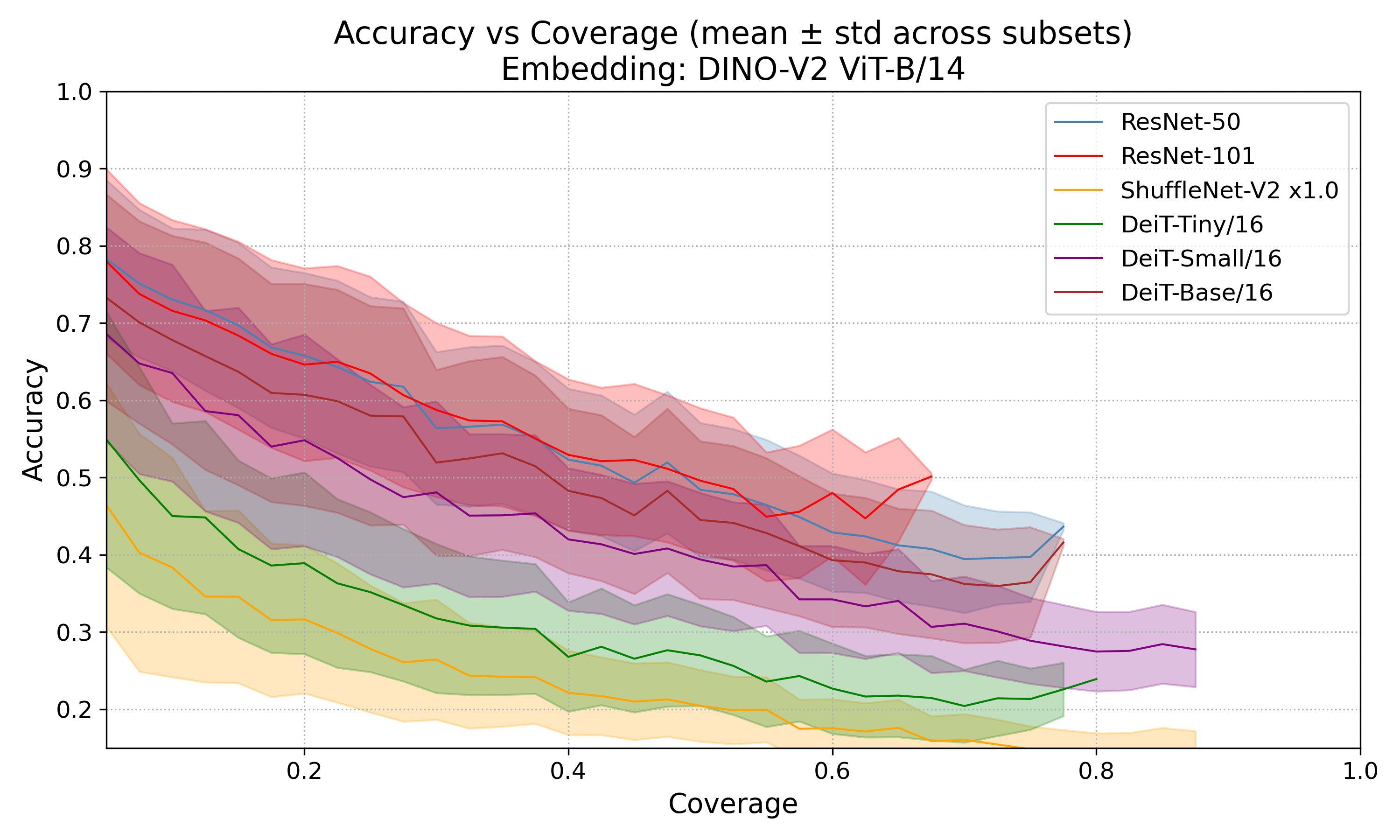}
      \caption{Confidence Curves for External Dataset. }
    \label{fig:fig_right}
  \end{subfigure}

   \caption{Accuracy vs. coverage (confidence curve) for the best embedding model ( DINO-V2 ViT-B/14 ).}
  \label{fig:two_model_groups}
\end{figure*}
}

\begin{table*}[b]
\centering
\caption{$N$ and best \textit{Normalized Confidence Gain} for each classification model across embedding models.}
\scriptsize
\begin{tabular}{lcccccc}
\toprule
\textbf{Classification Model} &
\textbf{DINO-V1 ViT-B/16} &
\textbf{DINO-V1 ViT-B/8} &
\textbf{DINO-V1 ViT-S/16} &
\textbf{DINO-V2 ViT-B/14} &
\textbf{DINO-V2 ViT-S/14} &
\textbf{MobileNet-V2} \\
\midrule
deit\_base\_patch16\_224       & 1 / 0.398 & 1 / 0.441 & 1 / 0.368 & 24 / \textbf{0.480} & 4 / 0.419 & 1 / 0.194 \\
deit\_small\_patch16\_224      & 1 / 0.408 & 1 / 0.449 & 1 / 0.379 & 7 / \textbf{0.472}  & 2 / 0.427 & 1 / 0.199 \\
deit\_tiny\_patch16\_224       & 1 / 0.402 & 1 / 0.433 & 1 / 0.379 & 23 / \textbf{0.442} & 4 / 0.406 & 1 / 0.218 \\
resnet101                      & 1 / 0.394 & 1 / 0.434 & 1 / 0.364 & 60 / \textbf{0.477} & 4 / 0.417 & 1 / 0.195 \\
resnet50                       & 1 / 0.407 & 1 / 0.448 & 1 / 0.380 & 24 / \textbf{0.487} & 2 / 0.426 & 1 / 0.209 \\
shufflenet\_v2\_x1\_0          & 1 / 0.418 & 1 / \textbf{0.436} & 1 / 0.401 & 7 / 0.417  & 4 / 0.395 & 1 / 0.244 \\
\bottomrule
\end{tabular}

\label{tab:best_n_integrals}
\end{table*}

For clarity, we now detail the procedure step by step using one specific classification model as an example:
    \paragraph{Step 1} For each image in the original training dataset ImageNet-1K (let us call this set the Base Set and the images in the set the Base Images), we calculate the embeddings for the models presented in Table~\ref{tab:embedding_models} using cosine similarity, which is a commonly adopted metric. We store these embeddings in the Base Embedding Database. Although the Base Image Set serves as a training set for the models we investigated, we do not fine-tune their parameters. Therefore, the word 'training' could be misleading, and we used the term 'base'.
    
    \paragraph{Step 2} For each image in the train set (a subset of ImageNet-V2), we compute its embedding and retrieve the $K_{\max}$ nearest neighbors from the Base Embedding Database. We refer to this collection as the train set. The value of $K_{\max}$ is chosen to be sufficiently large so that increasing it further does not change the outcome of the analysis.   The term 'train' emphasizes that this dataset is used to tune the parameters $N$ and $L$, which maximize the performance of the algorithm. These parameters are then kept fixed during the evaluation on the test sets. In addition, for each image in the train image set, we calculate and store the predictions of the classification model without any filtering for reference.

\paragraph{Step 3} For each image in the training set, we compute confidence statistics using two
parameters: the distance threshold $L$ and the neighbor-count threshold $N$.
For a given embedding, if the number of neighbors within distance $L$ exceeds
$N$, we treat the model’s prediction for that image as confident; otherwise, the
prediction is considered uncertain and is excluded from evaluation. We then
compute the classification accuracy over the subset of images that pass this
confidence test. We refer to the fraction of accepted images as \textit{coverage}.
Because increasing $L$ enlarges the acceptance region in embedding space, the
theoretical coverage function is non–decreasing in $L$. However, when using
approximate nearest-neighbor (ANN) search and floating-point distance
comparisons, we observe small local decreases in empirical coverage.  We
use coverage as a proxy variable: we evaluate only the discrete coverage values
that actually occur in the adjusting split and map each such coverage value to
the corresponding effective $L$. The resulting confidence curve is indexed by coverage rather than by $L$ itself. 

More precisely, let $d_1(x) \le d_2(x) \le \dots \le d_{K_{\max}}(x)$ denote the
sorted distances to the nearest neighbors of sample $x$. For each value
of $N$, we generate a coverage-ordered sequence by thresholding $L$ at these
distances. The effective threshold $L$ inducing a desired coverage level $\tau$
is therefore defined implicitly as
\[
L^{*}(N, \tau) = \min\{ L : \text{coverage}(L; N) \ge \tau \},
\]
where coverage$(L; N)$ is the fraction of samples whose $N$ closest neighbors
lie within radius $L$. Thus, $L^*$ is determined automatically by sweeping
through the sorted neighbor distances and selecting the smallest radius that
achieves the target coverage. 

The resulting \textit{confidence curve} plots accuracy as a function of coverage
(and therefore implicitly of $L$). Examples of confidence curves for ResNet-101
and DINO-V2 ViT-B/14 at several values of $N$ are shown in
Figure~\ref{fig:base_plot}(a). Figure~\ref{fig:two_model_groups}(b) illustrates the
corresponding \textit{maximum confidence gain} and \textit{confidence gain} for the
same embedding/model pair with $N=1$. Each point on the curve corresponds to a
specific effective value of $L$ induced by the optimal coverage.

\paragraph{Step 4} We find the values of $N$ that maximize the \textit{NCG} (the ratio between the \textit{Maximum Confidence Gain} and the \textit{Confidence Gain}) for the specific Embedding Model (to be discussed further). We use the optimal values of neighbors $N^*$ and the threshold value $L^*$ for the test sets.

\paragraph{Confidence Gains.}
We define the \textit{Confidence Gain} as the area between:  
(i) the confidence curve (with extrapolation if needed),  
(ii) the vertical line $\text{coverage}=0.1$,  
(iii) the vertical line $\text{coverage}=1$, and  
(iv) the horizontal line $\text{accuracy}=acc_b$,  
where $acc_b$ denotes the baseline accuracy obtained when all data samples are used (i.e., coverage = 1). This baseline depends solely on the classifier and dataset and is independent of the embedding model. Since the confidence curve approaches $acc_b$ as coverage approaches 1, the area can be computed directly using standard numerical integration (\texttt{Scikit-Learn}~\cite{scikit-learn}).

We define the \textit{Maximum Confidence Gain} as the area of the rectangle bounded by
$x=0.1$, $x=1.0$, $y=1.0$, and $y=acc_b$.
This value represents a theoretical upper bound on the achievable \textit{Confidence Gain}.
Consequently, we introduce the \textit{Normalized Confidence Gain} as the ratio between the
\textit{Confidence Gain} and its maximum possible value.
This normalized quantity lies in the interval $[0,1]$, where $1$ indicates an ideal confidence
curve and $0$ indicates no gain over the baseline.

Figure~\ref{fig:base_plot}(a) shows an example confidence curve for the ResNet-101 / DINO-V2 ViT-B/14 pair, although the same procedure applies to all model–embedding combinations. The distribution of points along the curve is uneven: at high coverage values, points become sparse, while at low coverage they are densely packed. This occurs because low coverage corresponds to highly confident predictions, which appear more frequently. To compensate for the lack of points near full coverage (i.e., near $\text{coverage}=1$), we linearly extrapolate the curve using the 20 points with the highest coverage values. We also observe that for very small coverage values (below 0.1), the curve becomes unstable and noisy. For robustness, we therefore restrict the computation of the \textit{Confidence Gain} to the interval $\text{coverage}\in[0.1, 1]$, where the accuracy varies smoothly and monotonically.

\begin{table*}[b]
\caption{The Normalized Confidence Gain for different pairs of embedding/classification models and for the combination of embedding models. Benchmark accuracy is shown for reference. All values are mean ± std across three label subsets for internal and external test sets.}
\centering
\tiny
\renewcommand{\arraystretch}{1.25}

\begin{tabular}{lccccccccc}
\toprule
\textbf{Set} & \textbf{Classification Model} & \textbf{Benchmark Acc.} & \textbf{DINO-V1 ViT-B/16} & \textbf{DINO-V1 ViT-B/8} & \textbf{DINO-V1 ViT-S/16} & \textbf{DINO-V2 ViT-B/14} & \textbf{DINO-V2 ViT-S/14} & \textbf{MobileNet-V2} & \textbf{Combination} \\
\midrule
\multirow{6}{*}{\rotatebox[origin=c]{90}{\textbf{internal test}}}
& deit\_base\_patch16\_224 & 0.696 $\pm$ 0.014 & 0.390 $\pm$ 0.052 & 0.414 $\pm$ 0.048 & 0.357 $\pm$ 0.035 & \textbf{0.449 $\pm$ 0.077} & 0.404 $\pm$ 0.046 & 0.194 $\pm$ 0.046 & 0.361 $\pm$ 0.058 \\
& deit\_small\_patch16\_224 & 0.666 $\pm$ 0.019 & 0.376 $\pm$ 0.068 & 0.403 $\pm$ 0.071 & 0.343 $\pm$ 0.062 & \textbf{0.431 $\pm$ 0.076} & 0.376 $\pm$ 0.058 & 0.189 $\pm$ 0.058 & 0.320 $\pm$ 0.060 \\
& deit\_tiny\_patch16\_224 & 0.582 $\pm$ 0.037 & 0.376 $\pm$ 0.076 & 0.388 $\pm$ 0.079 & 0.353 $\pm$ 0.078 & \textbf{0.408 $\pm$ 0.070} & 0.380 $\pm$ 0.067 & 0.202 $\pm$ 0.069 & 0.366 $\pm$ 0.067 \\
& resnet101 & 0.687 $\pm$ 0.025 & 0.345 $\pm$ 0.074 & 0.369 $\pm$ 0.066 & 0.308 $\pm$ 0.072 & \textbf{0.410 $\pm$ 0.106} & 0.366 $\pm$ 0.058 & 0.169 $\pm$ 0.065 & 0.316 $\pm$ 0.071 \\
& resnet50 & 0.676 $\pm$ 0.014 & 0.377 $\pm$ 0.050 & 0.408 $\pm$ 0.049 & 0.347 $\pm$ 0.050 & \textbf{0.446 $\pm$ 0.075} & 0.398 $\pm$ 0.044 & 0.189 $\pm$ 0.068 & 0.344 $\pm$ 0.061 \\
& shufflenet\_v2\_x1\_0 & 0.538 $\pm$ 0.022 & 0.363 $\pm$ 0.056 & 0.374 $\pm$ 0.063 & 0.345 $\pm$ 0.056 & \textbf{0.379 $\pm$ 0.053} & 0.357 $\pm$ 0.046 & 0.220 $\pm$ 0.050 & 0.351 $\pm$ 0.046 \\
\bottomrule
\end{tabular}

\vspace{1em}

\begin{tabular}{lccccccccc}
\toprule
\textbf{Set} & \textbf{Classification Model} & \textbf{Benchmark Acc.} & \textbf{DINO-V1 ViT-B/16} & \textbf{DINO-V1 ViT-B/8} & \textbf{DINO-V1 ViT-S/16} & \textbf{DINO-V2 ViT-B/14} & \textbf{DINO-V2 ViT-S/14} & \textbf{MobileNet-V2} & \textbf{Combinations} \\
\midrule
\multirow{6}{*}{\rotatebox[origin=c]{90}{\textbf{external test}}}
& deit\_base\_patch16\_224 & 0.269 $\pm$ 0.059 & 0.079 $\pm$ 0.109 & 0.132 $\pm$ 0.120 & 0.059 $\pm$ 0.107 & \textbf{0.214 $\pm$ 0.153} & 0.189 $\pm$ 0.149 & 0.044 $\pm$ 0.084 & 0.141 $\pm$ 0.130 \\
& deit\_small\_patch16\_224 & 0.229 $\pm$ 0.045 & 0.072 $\pm$ 0.092 & 0.120 $\pm$ 0.099 & 0.055 $\pm$ 0.082 & \textbf{0.188 $\pm$ 0.125} & 0.170 $\pm$ 0.119 & 0.037 $\pm$ 0.063 & 0.086 $\pm$ 0.087 \\
& deit\_tiny\_patch16\_224 & 0.152 $\pm$ 0.041 & 0.073 $\pm$ 0.077 & 0.102 $\pm$ 0.085 & 0.058 $\pm$ 0.072 & 0.119 $\pm$ 0.092 & \textbf{0.119 $\pm$ 0.093} & 0.040 $\pm$ 0.057 & 0.106 $\pm$ 0.090 \\
& resnet101 & 0.304 $\pm$ 0.057 & 0.069 $\pm$ 0.112 & 0.131 $\pm$ 0.112 & 0.049 $\pm$ 0.101 & \textbf{0.238 $\pm$ 0.144} & 0.211 $\pm$ 0.144 & 0.036 $\pm$ 0.080 & 0.147 $\pm$ 0.123 \\
& resnet50 & 0.293 $\pm$ 0.056 & 0.069 $\pm$ 0.102 & 0.127 $\pm$ 0.106 & 0.047 $\pm$ 0.101 & \textbf{0.243 $\pm$ 0.130} & 0.211 $\pm$ 0.135 & 0.035 $\pm$ 0.082 & 0.136 $\pm$ 0.114 \\
& shufflenet\_v2\_x1\_0 & 0.117 $\pm$ 0.027 & 0.062 $\pm$ 0.064 & 0.079 $\pm$ 0.065 & 0.047 $\pm$ 0.061 & \textbf{0.096 $\pm$ 0.069} & 0.090 $\pm$ 0.068 & 0.038 $\pm$ 0.048 & 0.083 $\pm$ 0.066 \\
\bottomrule
\end{tabular}
\label{tab:confidence_gain}
\end{table*}
\section{Results}
\begin{figure*}[t]
  \centering

  \begin{subfigure}{0.3\textwidth}
    \includegraphics[width=\linewidth]{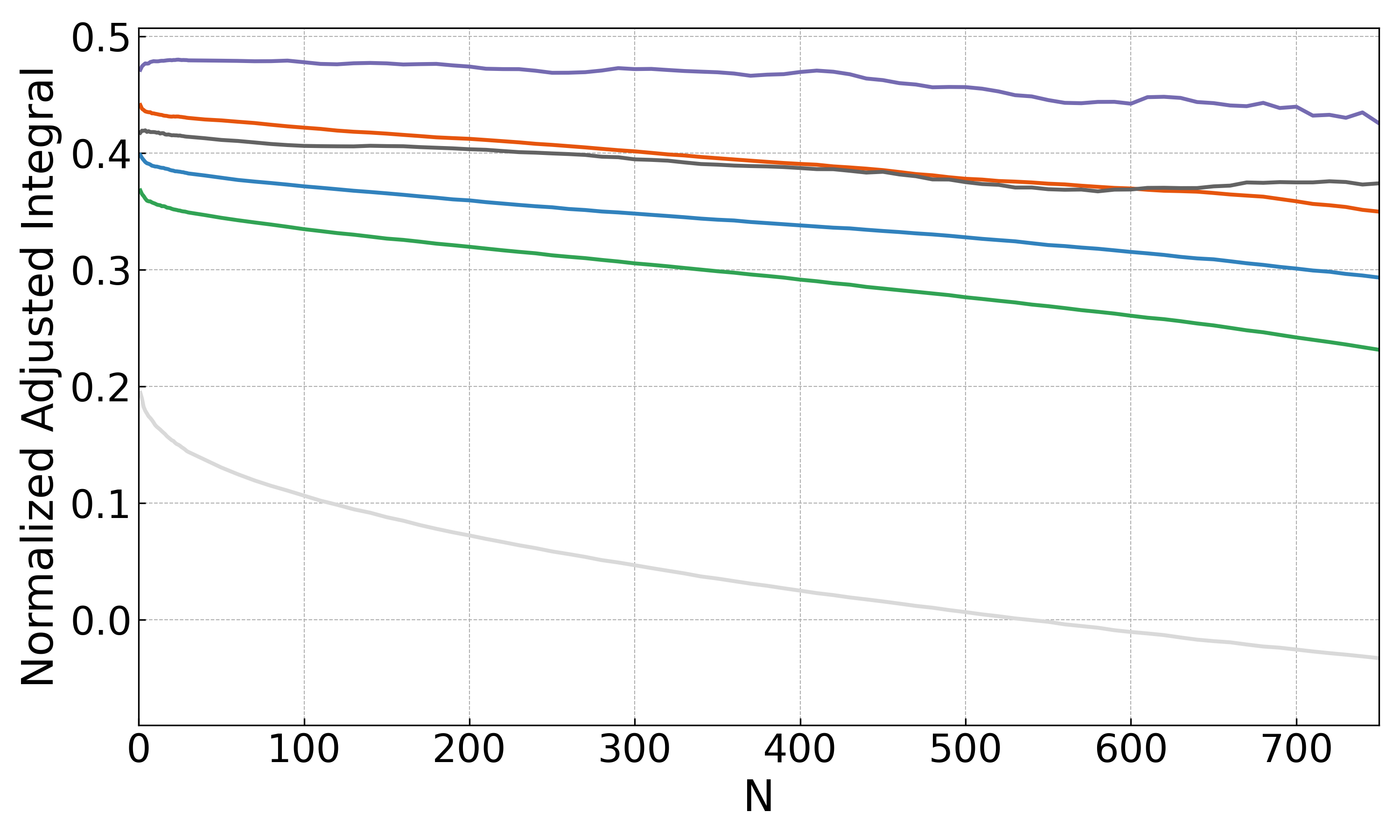}
    \caption{DeiT-Base}
  \end{subfigure}
  \hfill
  \begin{subfigure}{0.3\textwidth}
    \includegraphics[width=\linewidth]{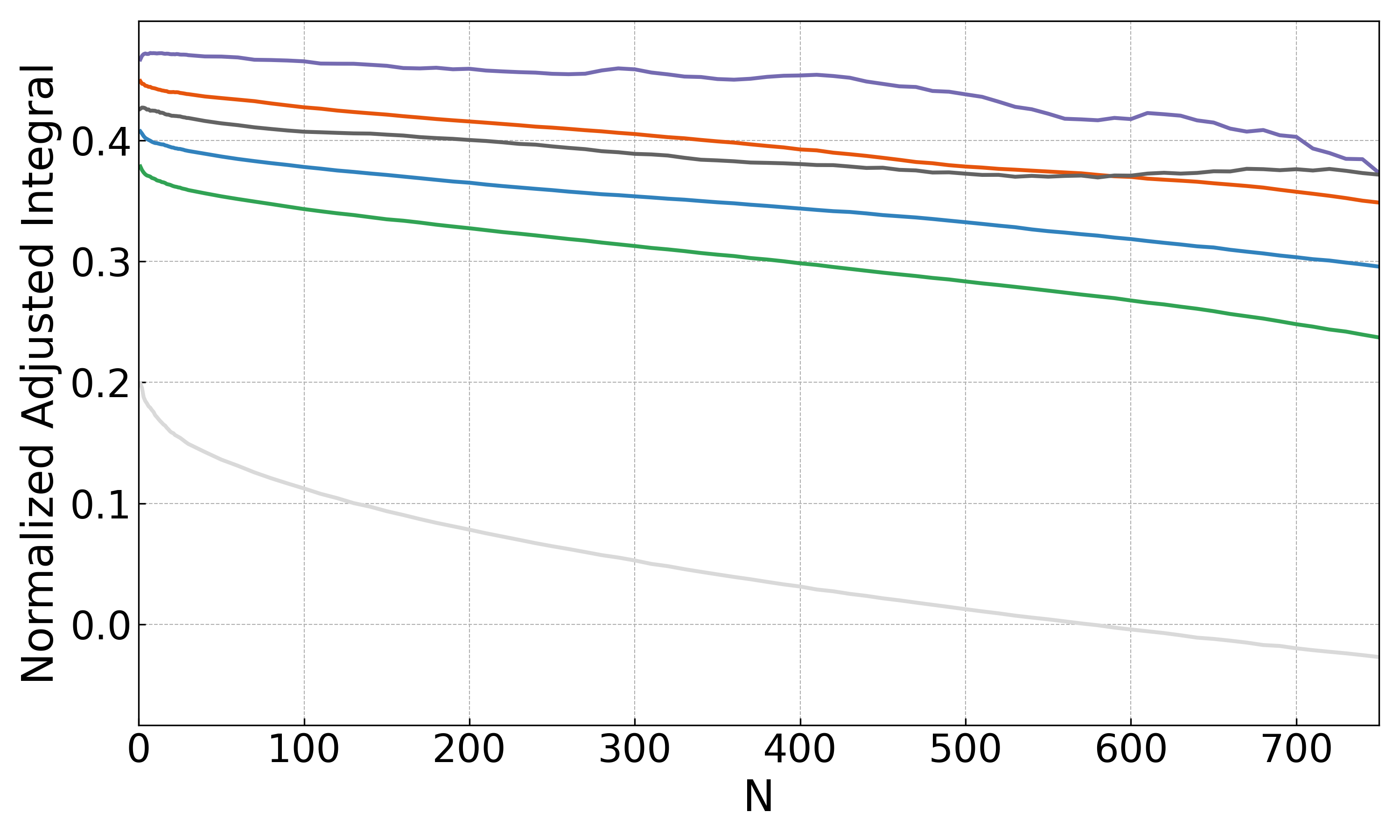}
    \caption{DeiT-Small}
  \end{subfigure}
  \hfill
  \begin{subfigure}{0.3\textwidth}
    \includegraphics[width=\linewidth]{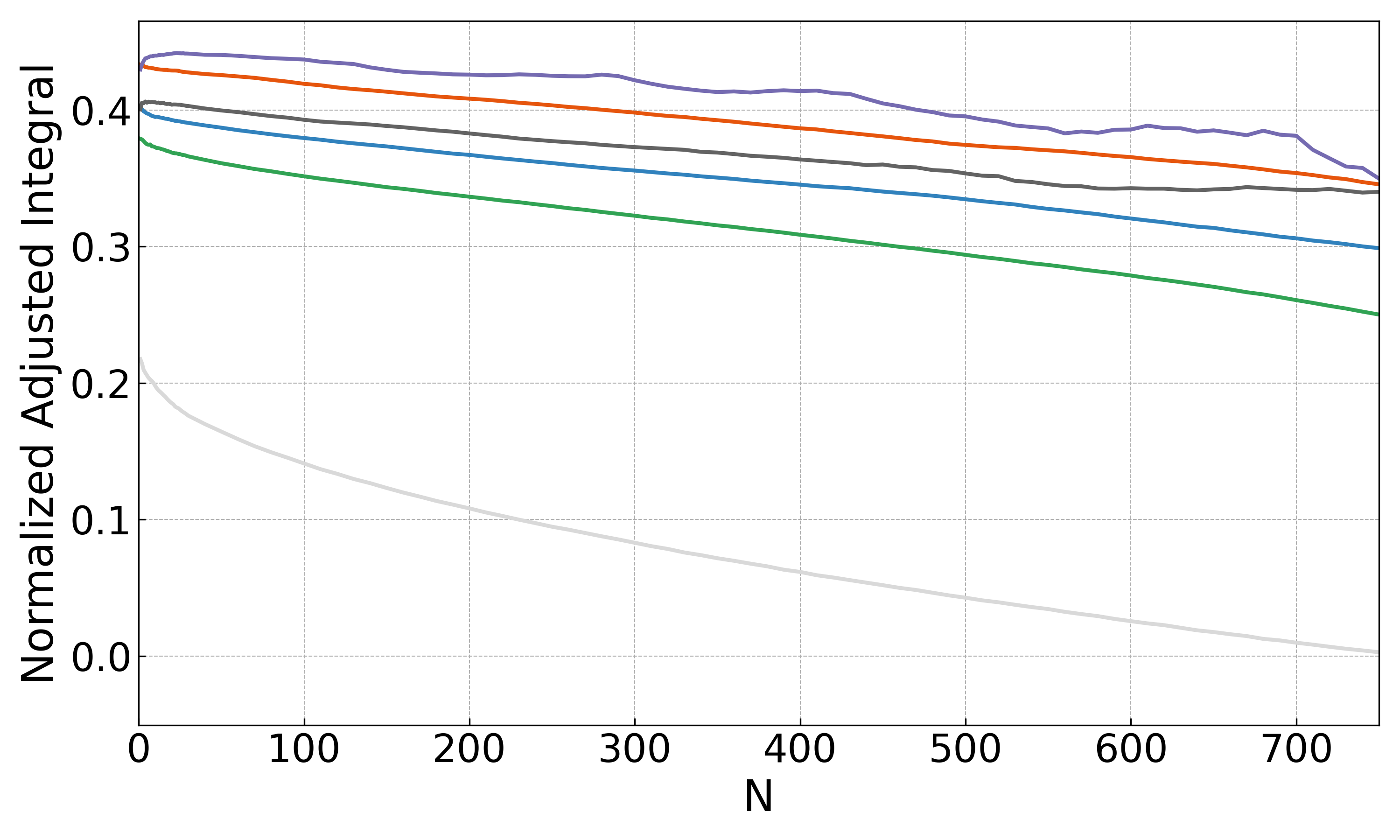}
    \caption{DeiT-Tiny}
  \end{subfigure}

  \vspace{1em}

  \begin{subfigure}{0.3\textwidth}
    \includegraphics[width=\linewidth]{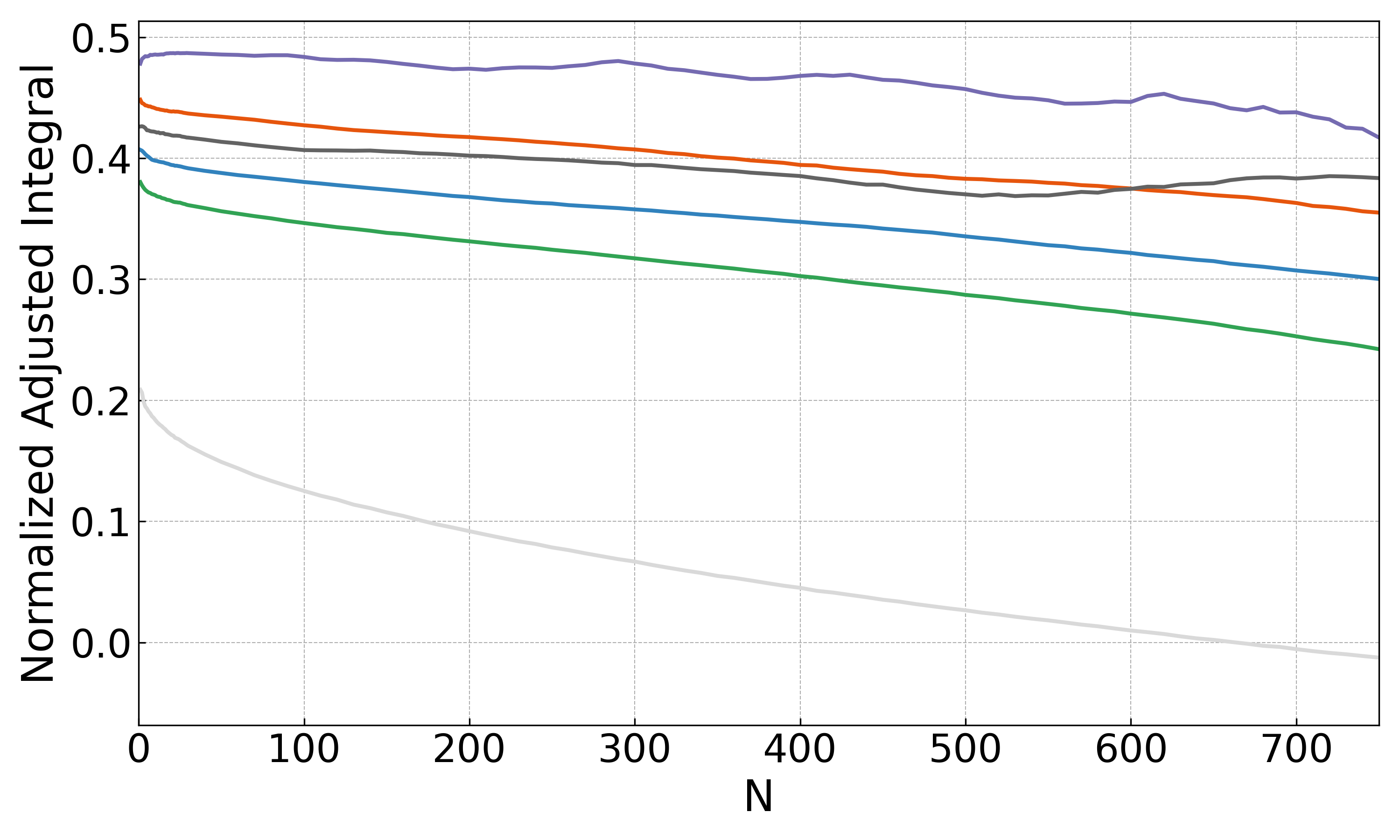}
    \caption{ResNet-50}
  \end{subfigure}
  \hfill
  \begin{subfigure}{0.3\textwidth}
    \includegraphics[width=\linewidth]{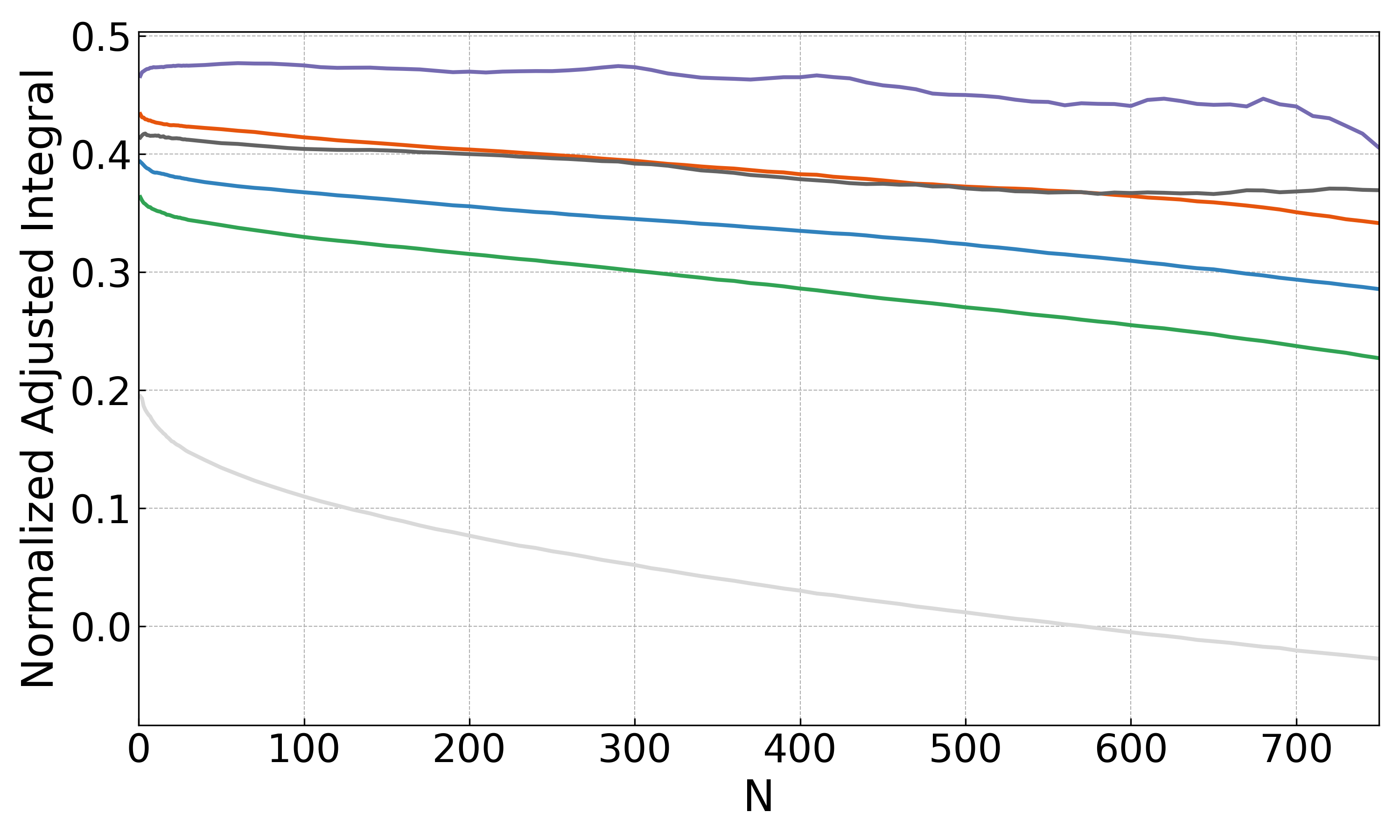}
    \caption{ResNet-101}
  \end{subfigure}
  \hfill
  \begin{subfigure}{0.3\textwidth}
    \includegraphics[width=\linewidth]{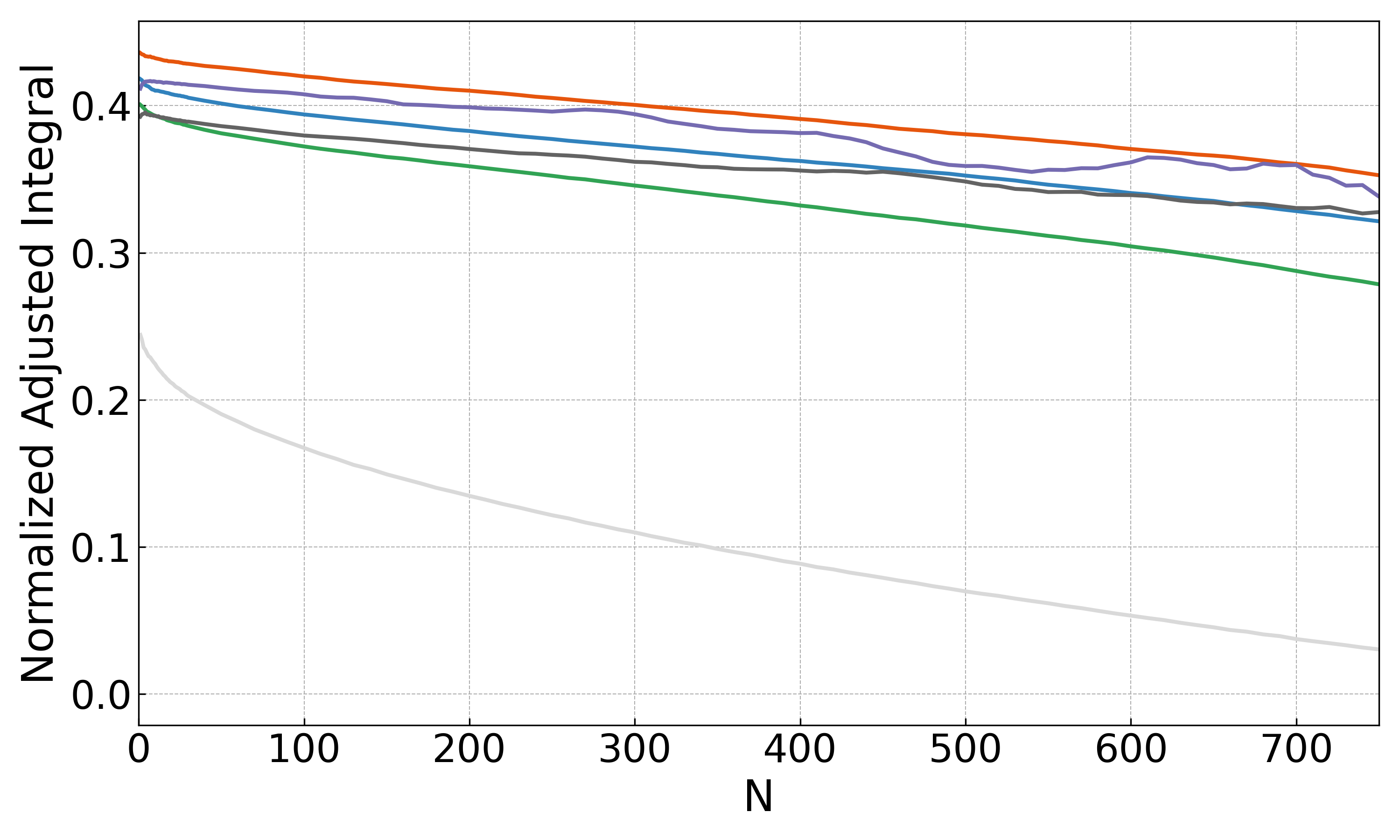}
    \caption{ShuffleNet-V2}
  \end{subfigure}

\vspace{1em}
\scalebox{0.6}{\includegraphics{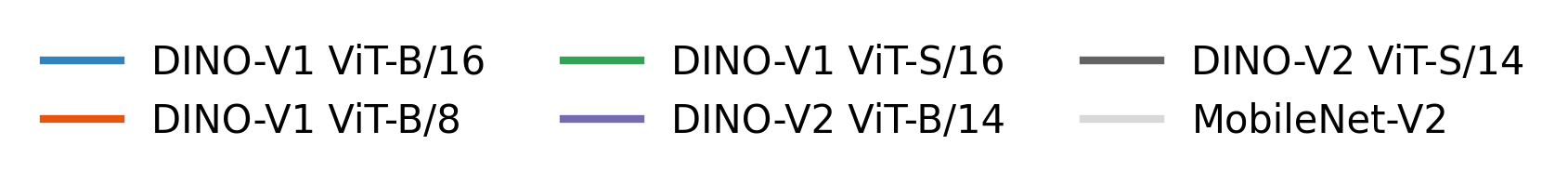}}

  \caption{Normalized Confidence Gain vs. number of neighbors $N$ for different classification models.}
  \label{fig:model_integrals}
\end{figure*}

\paragraph{ Sensitivity of the Confidence Gain.}
In Figure~\ref{fig:model_integrals}, the results of the adjustment (or training) are shown.  On the x-axis, the number of neighbors $N$ is presented, and on the y-axis, the calculated \textit{ Normalized Confidence Gain} is depicted.  We can observe that the highest \textit{Normalized Confidence Gain} value is achieved for all models except the ShuffleNet-V2 by the DINO-V2 ViT-B/14 embedding model, which is so far the strongest model among embedding models in terms of the number of parameters.  Interestingly, the second version of the DINO models outperforms the first version.  In Table \ref{tab:best_n_integrals}, the best parameter $N$ and the corresponding \textit{Normalized Confidence Gain} are presented.

For the evaluation pipeline, we select the number of parameters \(N\) that maximizes the \textit{NCG} for each pair of the classification model and embedding model. The corresponding value of \(N\) is then used to compute the metrics on both the internal test and external test sets. The resulting scores, along with their standard deviations (computed across the three subsets of each test set), are reported in Table~\ref{tab:confidence_gain}.

Across nearly all model combinations, the highest value of \textit{Normalized Confidence Gain} is obtained by DINO-v2 ViT-B/14. This behavior is expected and can be attributed to the model's higher capacity and the increased resolution of its input representations.

\begin{figure*}
    \centering
    \begin{subfigure}[t]{0.45\textwidth}
        \centering
        \includegraphics[width=\linewidth]{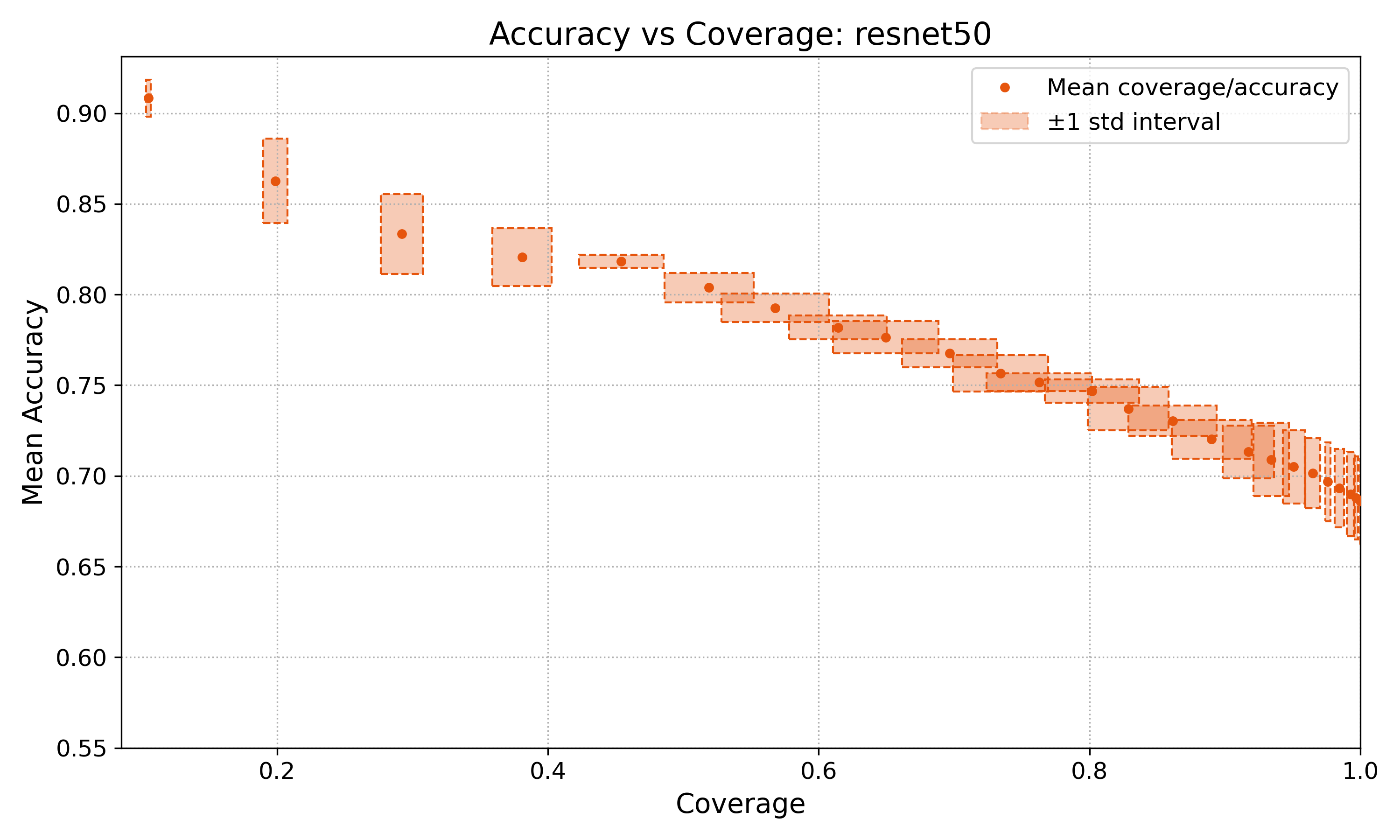}
        \caption{ResNet50 | Internal}
        \label{fig:resnet_internal}
    \end{subfigure}
    \hfill
    \begin{subfigure}[t]{0.45\textwidth}
        \centering
        \includegraphics[width=\linewidth]{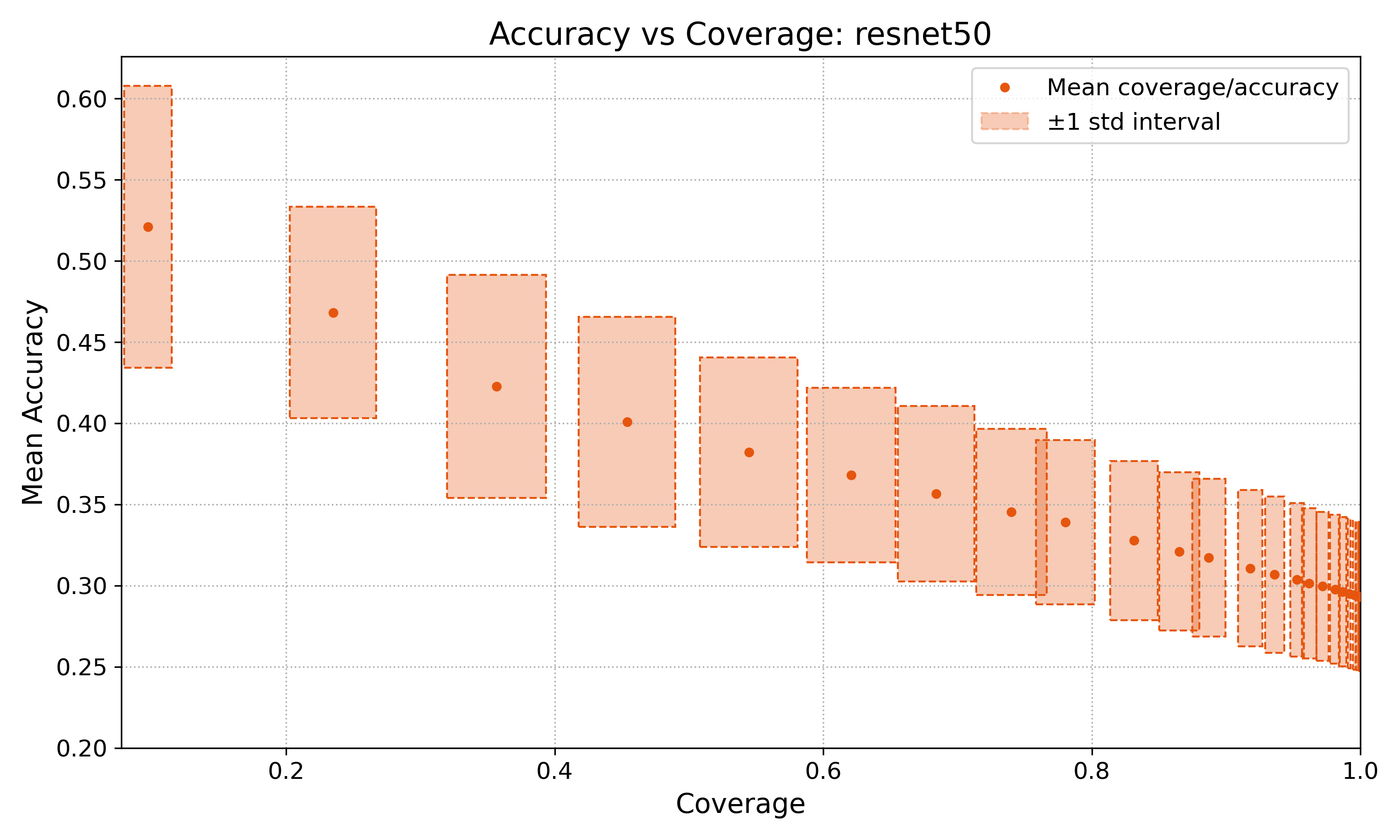}
        \caption{ResNet50 | External}
        \label{fig:resnet_external}
    \end{subfigure}

    \caption{Confidence curves (accuracy vs.\ total coverage) for the ResNet50 model evaluated on the internal (\textbf{left}) and external (\textbf{right}) datasets using ensemble of embedding models.}
    \label{fig:confidence_curve}
\end{figure*}

\paragraph{Ensemble confidence estimation.}
The next logical step would be to combine several embedding models (or use an approach similar to the ensemble of models, as in \cite{dietterich2000ensemble}), which has been done using greedy heuristics.  We fix the same coverage for each embedding model. We take the model with the highest \textit{Normalized Confidence Gain} and  make predictions. We accept a portion of the predictions and leave the rest to other models. Note that the total coverage for all embedding models is slightly higher than the parameter $coverage$ for individual models, but is not fixed. The results of the experiments (accuracy vs. coverage) are presented in Figure~\ref{fig:confidence_curve}. The \textit{Normalized Confidence Gain} for the combination of embedding models is shown in Table~\ref{tab:confidence_gain} in the column 'Combination'. The observed \textit{Confidence Gain} for the combination of embedding models is lower than that of the best-performing individual model.  In particular, at very low coverage values for the internal dataset, the accuracy of all models converges to similar levels. This suggests that the underlying data distribution plays a more decisive role than the intrinsic performance of the models. A plausible explanation is that, within the test datasets, certain images are located in close proximity (in the embedding space) to images from the training set. Consequently, under such conditions, the specific choice of the embedding model becomes less critical.  This is likely due  to the fact that the images in the external dataset (ObjectNet) exhibit a greater distributional shift relative to the original training set (ImageNet-1K) than those in the internal test set.

\paragraph{Choice of $K_{\max}$ and its effect on the confidence rule.}
In our method, confidence depends only on whether at least \(N^{*}\) neighbors fall within distance \(L^{*}\). Empirically, \(N^{*} \le 60\) for all model-embedding pairs, so the decision relies solely on the closest \(N^{*}\) neighbors. Under exact nearest-neighbor retrieval, retrieving \(K_{\max}\) neighbors and truncating to \(N^{*}\) is equivalent to retrieving only \(N^{*}\). Hence, for any \(K_{\max} \ge N^{*}\), the confidence decision is invariant to \(K_{\max}\); we set \(K_{\max}=1000\) as a conservative buffer. More generally, \(K_{\max}\) can be chosen empirically by tracking Normalized Confidence Gain (NCG) versus retrieval size. NCG typically rises or plateaus before declining as additional, noisy neighbors are included; the optimal \(K_{\max}\) lies near this onset. In our experiments, this stability region occurs well below 1000, further supporting \(K_{\max}=1000\) as a safe and efficient upper bound.

\paragraph{OOD score for embedding density filtering.}
To relate embedding density filtering to standard OOD evaluation metrics, such as
Mahalanobis distance, Energy score, and ODIN, we derive a continuous OOD score from our
embedding-density-based confidence criterion. This analysis is not intended as a direct
benchmark against logit-based OOD detectors, but rather as a diagnostic view of how
embedding-density-based confidence behaves under standard OOD evaluation protocols.

For a query image $x$, we retrieve its $N$ nearest neighbors
$\mathcal{N}_x = \{z_i\}_{i=1}^N$ from a fixed embedding database using cosine distance.
We define a continuous normalized embedding density score as
\[
s(x;N) = \frac{1}{N} \sum_{i=1}^{N} \bigl(1 - d_i\bigr),
\quad s(x;N) \in [0,1],
\]
where $d_i$ denotes the cosine distance between the query embedding $m(x)$ and its
$i$-th nearest neighbor. Higher values of $s(x;N)$ indicate dense, well-supported regions
of the training distribution, while lower values correspond to sparser regions in
embedding space.

Following standard OOD evaluation conventions, where larger scores indicate higher
out-of-distribution likelihood, we define the OOD score as
\[
\mathrm{OOD}(x;N) = 1 - s(x;N).
\]
Samples that lie in low-density regions of embedding space therefore, receive high OOD
scores, while samples well supported by the training data receive low OOD scores. The time complexity of the algorithm is described in
Appendix~\ref{app:time_complexity}. Unlike most logit-based OOD metrics, the proposed approach allows the neighborhood size $N$
and distance threshold $L$ to be investigated cheaply as a postprocessing step. In our experiments, we evaluate this score over neighborhood sizes
$N \in \{1,5,10,20,30,50,75,100\}$ and distance thresholds
$L \in [0.05,\,0.6]$ with a step size of $0.05$.

\paragraph{Mahalanobis distance.}
We employ the Mahalanobis distance method of \cite{lee2018simple}. In this approach, in-distribution samples are
modeled as a single multivariate Gaussian in  a feature space of a pretrained model (classifier backbone). The mean vector and covariance matrix are estimated from a subset of
ImageNet-1K training images, and a shrinkage term is added to the covariance
matrix for numerical stability. For a query image $x$ with feature representation
$f(x)$, the OOD score is given by
\[
\mathrm{OOD}_{\mathrm{Mah}}(x)
=
(f(x) - \mu)^{\top} \Sigma^{-1} (f(x) - \mu),
\]
where larger values indicate a higher likelihood of being out-of-distribution.
We follow the standard evaluation protocol of \cite{lee2018simple} without
additional tuning.

\paragraph{Energy-based OOD detection.}
We additionally compare against the Energy-based OOD detection method proposed
by \cite{liu2020energy}. This approach derives an OOD score directly from the
classifier logits without requiring access to intermediate features or
additional density estimation. Given the logits $f(x)$ of a classifier and a
temperature parameter $T$, the Energy score is defined as
\[
\mathrm{OOD}_{\mathrm{Energy}}(x)
=
- T \log \sum_{k} \exp\!\left( \frac{f_k(x)}{T} \right),
\]
where lower energy values indicate in-distribution samples and higher values
indicate out-of-distribution samples.

In our experiments, we evaluate the Energy score over a discrete set of
temperature values
\[
T \in \{0.5,\;1.0,\;5.0\},
\]
and select the optimal temperature on the adjustment split based on AUROC,
following the standard practice of \cite{liu2020energy}. 

\paragraph{ODIN.}
We further compare against the ODIN method proposed by \cite{liang2018odin},
which improves OOD detection by combining temperature scaling of logits with a
small, input-dependent perturbation. Given classifier logits $f(x)$, ODIN first
applies temperature scaling with parameter $T$ and then computes the maximum
softmax probability as the confidence score.

Specifically, for a given temperature $T$, an input perturbation is generated by
taking a single gradient step that minimizes the cross-entropy loss with respect
to the predicted class. The perturbed input is given by
\[
x_{\mathrm{adv}} = x - \epsilon \,\mathrm{sign}\!\left(
\nabla_x \mathcal{L}\big(f(x)/T, \hat{y}\big)
\right),
\]
where $\hat{y}$ is the predicted class and $\epsilon$ controls the perturbation
magnitude. The final ODIN score is defined as the maximum softmax probability
computed from the perturbed input and temperature-scaled logits.

In our experiments, we evaluate ODIN over a grid of temperature and perturbation
parameters,
\[
T \in \{1.0,\;10.0,\;100.0\},
\qquad
\epsilon \in \{0.0,\;0.001,\;0.002\},
\]
and select the optimal parameter pair $(T,\epsilon)$ on the adjustment split
based on AUROC, following the standard protocol of \cite{liang2018odin}.

\paragraph{Comparison with different OOD methods.}
For a fair and robust evaluation, both the \textit{Internal} and \textit{External} datasets were partitioned into three mutually exclusive subsets based on class labels, such that each label appeared in exactly one subset. OOD detection metrics were computed independently on each subset, and the reported results correspond to the mean and standard deviation across these three splits. We report standard OOD evaluation metrics, including AUROC, AUPR$_{\text{Out}}$, AUPR$_{\text{In}}$, and FPR95. For methods requiring hyperparameters, such as Energy-based scoring, the optimal parameters (e.g., temperature $T$) were selected based on validation performance and are explicitly reported in Table~\ref{tab:ood_internal_external}. Methods without tunable parameters are marked accordingly. For a comparison of computational efficiency and runtime, we refer the reader to Appendix~\ref{app:time_comparison}.

\begin{table*}[t]
\caption{OOD detection performance on Internal and External datasets}
\centering
\tiny
\setlength{\tabcolsep}{4pt}
\renewcommand{\arraystretch}{1.25}

\begin{tabular}{p{1.2cm} l l c c c c c c}
\toprule
\textbf{Set} & \textbf{Backbone} & \textbf{Method} & \textbf{AUROC} & \textbf{AUPR$_{Out}$} & \textbf{AUPR$_{In}$} & \textbf{FPR95} & \textbf{Best Acc.} & \textbf{Best parameters} \\
\midrule

\multirow{7}{*}{\rotatebox[origin=c]{90}{\textbf{Internal}}}

& \multirow{3}{*}{ResNet101}
  & Energy
  & 0.5915 $\pm$ 0.0046
  & 0.0010 $\pm$ 0.0000*
  & 0.9995 $\pm$ 0.0000*
  & 0.9251 $\pm$ 0.0150
  & 0.9994 $\pm$ 0.0000*
  & T=0.5 \\
& & Mahalanobis
  & 0.5680 $\pm$ 0.0110
  & 0.0233 $\pm$ 0.0008
  & 0.9859 $\pm$ 0.0008
  & 0.9432 $\pm$ 0.0073
  & 0.9836 $\pm$ 0.0001
  & None \\
& & ODIN
  & 0.6230 $\pm$ 0.0110
  & 0.0019 $\pm$ 0.0001
  & 0.9994 $\pm$ 0.0000*
  & 0.9191 $\pm$ 0.0103
  & 0.9992 $\pm$ 0.0000*
  & T=1.0,\ $\epsilon$=0.0 \\
\noalign{\vskip 0.5em}
& \multirow{3}{*}{DeiT-Tiny}
  & Energy
  & 0.5858 $\pm$ 0.0065
  & 0.0009 $\pm$ 0.0000*
  & 0.9995 $\pm$ 0.0000*
  & 0.9263 $\pm$ 0.0075
  & 0.9994 $\pm$ 0.0000*
  & T=0.5 \\
& & Mahalanobis
  & 0.5850 $\pm$ 0.0130
  & 0.0269 $\pm$ 0.0021
  & 0.9874 $\pm$ 0.0007
  & 0.9157 $\pm$ 0.0134
  & 0.9836 $\pm$ 0.0001
  & None \\
& & ODIN
  & 0.5878 $\pm$ 0.0053
  & 0.0009 $\pm$ 0.0000*
  & 0.9995 $\pm$ 0.0000*
  & 0.9249 $\pm$ 0.0041
  & 0.9994 $\pm$ 0.0000*
  & T=1.0,\ $\epsilon$=0.0 \\
\noalign{\vskip 0.5em}

& \textbf{DINO-V2 ViT-B/14}
  & \textbf{Embedding Density (Ours)}
  & 0.5867 $\pm$ 0.0044
  & 0.0008 $\pm$ 0.0000*
  & 0.9995 $\pm$ 0.0000*
  & 0.8981 $\pm$ 0.0050
  & 0.9994 $\pm$ 0.0000*
  & N=1,\ L=0.05 \\
\midrule

\multirow{7}{*}{\rotatebox[origin=c]{90}{\textbf{External}}}

& \multirow{3}{*}{ResNet101}
  & Energy
  & 0.9001 $\pm$ 0.0060
  & 0.1300 $\pm$ 0.0060
  & 0.9983 $\pm$ 0.0002
  & 0.4105 $\pm$ 0.0363
  & 0.9871 $\pm$ 0.0000*
  & T=0.5 \\
& & Mahalanobis
  & 0.8950 $\pm$ 0.0058
  & 0.6538 $\pm$ 0.0090
  & 0.9646 $\pm$ 0.0028
  & 0.3150 $\pm$ 0.0262
  & 0.8384 $\pm$ 0.0041
  & None \\
& & ODIN
  & 0.8891 $\pm$ 0.0069
  & 0.2161 $\pm$ 0.0068
  & 0.9971 $\pm$ 0.0003
  & 0.5893 $\pm$ 0.0500
  & 0.9849 $\pm$ 0.0002
  & T=1.0,\ $\epsilon$=0.0 \\

\noalign{\vskip 0.5em}

& \multirow{3}{*}{DeiT-Tiny}
  & Energy
  & 0.7979 $\pm$ 0.1368
  & 0.1304 $\pm$ 0.0208
  & 0.9861 $\pm$ 0.0170
  & 0.6120 $\pm$ 0.2275
  & 0.9752 $\pm$ 0.0168
  & T=0.5 \\
& & Mahalanobis
  & 0.7243 $\pm$ 0.0003
  & 0.3779 $\pm$ 0.0001
  & 0.8981 $\pm$ 0.0002
  & 0.6344 $\pm$ 0.0053
  & 0.7489 $\pm$ 0.0005
  & None \\
& & ODIN
  & 0.8880 $\pm$ 0.0047
  & 0.1202 $\pm$ 0.0026
  & 0.9982 $\pm$ 0.0002
  & 0.4577 $\pm$ 0.0261
  & 0.9880 $\pm$ 0.0002
  & T=10.0,\ $\epsilon$=0.0 \\

\noalign{\vskip 0.5em}

& \textbf{DINO-V2 ViT-B/14}
  & \textbf{Embedding Density (Ours)}
  & 0.8512 $\pm$ 0.0005
  & 0.0485 $\pm$ 0.0001
  & 0.9975 $\pm$ 0.0000*
  & 0.4426 $\pm$ 0.0045
  & 0.9871 $\pm$ 0.0000*
  & N=1,\ L=0.05 \\

\bottomrule
\end{tabular}

\label{tab:ood_internal_external}

\vspace{0.3em}
{\scriptsize $^*$ Standard deviation $< 10^{-4}$.}
\end{table*}

For OOD evaluation, samples from ImageNet-1K  are treated as
in-distribution, while samples from ObjectNet and ImageNet-V2 (validation subset) are treated as out-of-distribution.
The OOD score $\mathrm{OOD}(x)$ is computed using the optimal parameters
$(N^{\ast}, L^{\ast})$ selected on the adjustment split. Standard OOD metrics,
including AUROC, AUPR$_{\mathrm{Out}}$, AUPR$_{\mathrm{In}}$, and FPR95, are then
computed by sweeping a threshold over $\mathrm{OOD}(x)$, following the same
protocol used for the baseline OOD methods.

Table~\ref{tab:ood_internal_external} reports OOD detection performance on Internal
and External datasets. In the Internal setting, all methods—including Energy,
Mahalanobis, ODIN, and the proposed embedding density score—exhibit near-chance
AUROC ($\approx 0.58$--$0.59$) and very high $\mathrm{FPR}_{95}$, indicating substantial overlap between
in-distribution and internally shifted samples. In this regime, the embedding
density score closely matches logit-based baselines despite relying on a
deliberately simple density-based scoring function \( S(x) \) defined on
nearest-neighbor distances in embedding space; notably, using only the single
nearest neighbor ($N=1$) is sufficient to achieve this performance.

In contrast, performance improves markedly on the External dataset, where the
distribution shift is stronger. Energy and ODIN achieve the highest AUROC
($\approx 0.90$), benefiting from access to classifier logits and calibration, while the
proposed method attains a competitive AUROC of 0.85 under strictly weaker
assumptions.

Overall, the embedding density score follows standard OOD behavior: it degrades
when representation overlap is high and improves under larger distribution
shifts, with the remaining gap highlighting the additional discriminative signal
provided by classifier-specific information beyond embedding geometry alone.

\paragraph{Applicability of the comparison of OOD methods.}

OOD detection methods differ fundamentally in the information they require and the architectures on which they operate. Logit-based approaches such as Energy and ODIN rely on classifier outputs (and, in some cases, input gradients) and are therefore applicable only to supervised models with task-specific heads. In contrast, the proposed embedding density score operates solely on frozen feature embeddings and does not require classifiers, logits, or gradients. Consequently, methods are evaluated under different backbones: Energy, Mahalanobis, and ODIN use supervised classifiers, while embedding density uses a self-supervised DINO-v2 ViT backbone. Enforcing a shared representation would require auxiliary classifiers or probes, altering the assumptions of several methods and violating the embedding-only setting. Accordingly, each method is evaluated within its native regime, and the comparison \textbf{should not be interpreted as a representation-controlled benchmark}. Rather, it highlights qualitative differences between classifier-dependent and purely geometric OOD signals.

\section{Discussion}

From the experimental results, we infer that a more sophisticated embedding model is more likely to have a better Confidence Curve (or higher \textit{NCG}). By ‘sophisticated,’ we mean a larger number of dimensions, higher input image resolution, and a richer dataset on which the model was trained. In particular, the resolutions of the images and the capacity of the embedding model significantly determine the performance.  

We observe that after training, most of the models achieved the highest \textit{NCG} with $N=1$. Only the best-performing embedding model achieved the maximum value at significantly larger values of $N$ ($N>1$). We believe that the more complicated the model, the larger the optimal value of $N$. 
To clarify whether the superior performance of DINO-V2 ViT-B/14 arises primarily from its increased embedding capacity or from differences in pretraining data, we performed a controlled analysis comparing models with matched dimensionality and similar architectural scale where possible (e.g., DINO-V1 ViT-B/16 vs. DINO-V2 ViT-B/14, both 768-dimensional embeddings). Across all classification models and test splits, DINO-V2 consistently achieved a higher NCG margin of 0.04–0.07 over its DINO-V1 counterpart at matched dimensionality, even when image resolution was held constant. In contrast, increasing embedding dimensionality within the same family (e.g., DINO-V1 ViT-S/16 → ViT-B/16 → ViT-B/8) produced smaller NCG improvements of 0.01–0.03. This suggests that the majority of the observed gain is attributable to the broader and more diverse pretraining corpus of DINO-V2 (142M curated images) rather than architectural scaling alone. This conclusion is further supported by the fact that MobileNet-V2, which has a higher-dimensional embedding (1000-D) but a much narrower pretraining distribution, performs significantly worse in NCG despite its larger embedding space. Overall, while embedding capacity contributes modestly to the effect, our controlled comparisons indicate that the dominant factor behind DINO-V2’s improved confidence behavior is the data distribution on which it is pretrained, rather than dimensionality or model size alone.

The external dataset (ObjectNet) demonstrates smaller absolute improvements, largely due to the substantially lower baseline accuracy of all classification models on this benchmark compared to the internal dataset. Because the \textit{NCG} metric is computed relative to each model’s own accuracy, a weaker baseline inherently limits absolute  achievable gain. 

Combining embeddings from multiple models offers a potential way to integrate diverse representation spaces; however, in our experiments, such combinations did not outperform the best individual embedding model in terms of \textit{NCG}. Unlike classical ensemble methods, where aggregating weaker models often yields gains, we did not observe a similar effect in this setting. This may be due to the simplicity of our greedy combination strategy, as more sophisticated approaches (e.g., per-sample model selection) could be more effective, as well as the limited diversity of pretraining data across the evaluated models, which constrains the potential for complementary information.

An important aspect of the reported OOD results is that the embedding density method employs an intentionally simple scoring function. In all experiments, the OOD score is derived from a straightforward aggregation of nearest-neighbor distances in embedding space, without any learned parameters, calibration, class conditioning, or access to classifier logits (only calibration of distance threshold $L$ and optimal number of neighbors $N$). Despite this simplicity, the resulting OOD performance closely matches that of logit-based baselines in the Internal setting and remains competitive on the External dataset. This suggests that a substantial portion of the OOD signal captured by more complex methods is already present in the local geometry of the representation space. At the same time, the remaining performance gap on the External benchmark could indicate the additional information contributed by classifier-specific mechanisms, rather than a deficiency of the underlying representation geometry itself.

An additional practical advantage of the proposed embedding density score for the OOD detection is that it enables detection without performing classification at inference time. Because the score depends solely on distances in a frozen embedding space, it can be computed for arbitrary inputs even when class labels, classifier heads, or calibrated logits are unavailable. In contrast, logit-based methods such as Energy and ODIN inherently require a trained classifier and access to its output scores, which restricts their applicability to supervised classification pipelines. This distinction makes embedding-density–based OOD detection particularly suitable for representation-centric settings, such as retrieval systems, self-supervised or foundation models, and scenarios where classification is not the primary objective.

While our experiments focus on image classification, the proposed embedding density framework may extend to other modalities, particularly NLP. A central challenge in this setting is the absence of a natural discretization of text: unlike images, text forms a continuous sequence, making the choice of window size and stride nontrivial. Although multi-scale windowing could capture semantic structure at different levels, dense sliding windows quickly become computationally prohibitive at corpus scale. Large language model training datasets contain hundreds of billions of tokens, rendering naive per-token embedding storage infeasible without substantial compression or aggregation. Addressing these challenges remains an important direction for future work.

\section{Conclusion}
We introduced a framework for estimating prediction confidence using distances in embedding space. Rather than modifying or retraining the classifier, our method identifies samples lying outside the training distribution. This leads to consistent improvements in effective accuracy while discarding a fraction of data. We further showed that a single embedding model is sufficient to filter unreliable predictions across diverse classifiers. Because confidence is tied to the geometry of the training manifold, the approach naturally reacts to distribution shift, providing an implicit mechanism for drift detection. Overall, our results demonstrate that leveraging training embeddings offers a practical and relatively lightweight alternative for confidence estimation in real-world settings where data distributions evolve over time. Code for the experiments is available at \href{https://github.com/maksimkazanskii/prior_hallucinations}{GitHub repository} (\url{https://github.com/maksimkazanskii/prior_hallucinations}).

\section{Use of Generative AI Tools}
Generative AI tools were used only for language editing, formatting, and minor code refactoring. All scientific reasoning, analysis, methodology, and conclusions are solely the responsibility of the authors.

\bibliographystyle{unsrt}   
\bibliography{main}

\clearpage
\appendix

\setlength{\parskip}{0pt} 

\clearpage
\setlength{\parskip}{0pt} 

\section{Appendix}

\subsection{Choosing the optimal values}

\paragraph{Search for optimal threshold parameters.}
The search for the optimal pair of threshold parameters $(L^{\ast}, N^{\ast})$
is performed over discrete grids of candidate distance thresholds
$\mathcal{L}$ and neighbor-count values $\mathcal{N}$. The distance threshold
grid is defined as
\[
\mathcal{L} = \{-0.20,\,-0.195,\,-0.190,\,\ldots,\,1.000\},
\]
constructed using a uniform step size of $0.005$, yielding 241 distinct
thresholds. This is valid for the cosine similarity metric; for other metrics
the potential grid is significantly different. The lower bound of $-0.20$ accommodates the slight
negative cosine similarities that may arise for some embedding models while
avoiding the computational expense of sweeping the full $[-1,1]$ range.

The search procedure consists of two stages. In the first stage, for each
$L \in \mathcal{L}$ and for each query sample, we compute how many of its
$K_{\max}$ retrieved nearest neighbors satisfy the similarity constraint
(similarity $> L$). This preprocessing step has time complexity
\[
O\!\left( N_{\text{train}}\, K_{\max}\, |\mathcal{L}| \right),
\]

In the second stage, the algorithm sweeps over all threshold pairs $(L, N)$ with
$L \in \mathcal{L}$ and $N \in \mathcal{N}$, evaluating coverage and accuracy
using the precomputed neighbor counts. This stage has time complexity
\[
O\!\left( N_{\text{train}}\, |\mathcal{L}|\, |\mathcal{N}| \right),
\]

\medskip

\textbf{Integral-based ranking of $N$.}
\label{app:integral}
For each fixed $N$, the smoothed coverage–accuracy curve is denoted by $s(x)$.
After extrapolating $s(x)$ to the full interval $[0.1,1.0]$, we compute the
integral score
\[
I(N) = \int_{0.1}^{1.0} s(x)\,dx.
\]
To account for model-dependent baseline accuracy $A_{\mathrm{bench}}$, we define
\[
I_{\mathrm{adj}}(N) = I(N) - 0.9\,A_{\mathrm{bench}},
\]
\[
I_{\mathrm{norm}}(N)
= \frac{I_{\mathrm{adj}}(N)}{0.9\,(1 - A_{\mathrm{bench}})}.
\]
The optimal neighbor-count parameter is then
\[
N^{\ast} = \arg\max_{N \in \mathcal{N}} I_{\mathrm{norm}}(N).
\]

\medskip

The computational cost of the integral evaluation step is modest.
$L$ would  be the number of coverage–accuracy points per curve. Spline
fitting and numerical integration
leading to a total of
\[
O(|\mathcal{N}|\,|\mathcal{L}|)
\]
This is negligible compared to the cost of
generating the underlying coverage and accuracy values.

\subsection{Computational complexity (storing and querying)}
\label{app:time_complexity}
The computational cost of the proposed method has two main components:
(i) storing embedding vectors for the base (training) dataset, and
(ii) retrieving nearest neighbors for each query.

 Let $N_{\text{base}}$ denote
the number of stored embeddings and $d$ the embedding dimensionality. Storing
one $d$-dimensional vector per sample results in a memory complexity of
\[
O(N_{\text{base}} d),
\]
which grows linearly with dataset size. For example, DINOv2-B/14 embeddings
($d=768$) require approximately 3~KB per image, corresponding to about 3.8~GB
for the ~1.28M images in ImageNet-1K.

However, the bottleneck is the computation of the embedding vector itself. For DINOv2-B, computing an embedding for a \emph{single image} (batch size = 1)
takes approximately $0.05\,\text{s}$, whereas storing the resulting embedding
vector in the database requires only $0.005\,\text{s}$. These measurements were
obtained without batch processing and therefore reflect the true per-image
latency of the pipeline.
A structured comparison of both computational and memory costs is presented in
Table~\ref{tab:storage_complexity}.

\begin{table}[h]
\centering
\caption{Computational and memory complexity for DINOv2-B and ChromaDB on CPU
(Mac M1 Pro). Values are shown per image and for the full ImageNet-1K dataset
($\sim$1.28M images).}

\setlength{\tabcolsep}{4pt} 
\renewcommand{\arraystretch}{0.9}

\scalebox{0.7}{
\begin{tabular}{lcc}
\toprule
\textbf{Category} & \textbf{Per image} & \textbf{Per dataset} \\
\midrule
\textbf{Computational complexity of storing and querying} & & \\
Encoding time        & 0.050 s  & 140,800 s (39.1 h) \\
Recording time       & 0.005 s  & 6,400 s (1.78 h) \\
\textbf{Total time}  & 0.055 s  & 147,200 s (40.9 h) \\
\midrule
\textbf{Memory complexity} & & \\
Storage per embedding  & $\sim$3 KB & --- \\
\textbf{Total storage} & ---  & $\sim$3.8 GB \\
\bottomrule
\end{tabular}
}
\label{tab:storage_complexity}
\end{table}

Let $N_{\text{base}}$ denote the number of stored embeddings. Querying an HNSW
approximate nearest-neighbor (ANN) index scales as
\[
O(\log N_{\text{base}} \cdot d),
\]
where $d$ is the embedding dimension, yielding sublinear retrieval time even for
databases with millions of entries.

In practice, ANN retrieval is not the dominant cost. The primary bottleneck is
feature extraction: computing a DINOv2-B/14 embedding for a single image takes
approximately $0.05\,\text{s}$, which dominates the total per-image runtime.
Consequently, overall evaluation time is determined mainly by the embedding
network forward pass rather than by nearest-neighbor search.

\begin{table}[b]
\centering
\footnotesize
\captionsetup{width=\linewidth}
\caption{Computational cost for DINOv2-B embeddings and ANN retrieval on CPU.
Values are reported per image and for $\sim$1.28M images (ImageNet-1K scale).}
\label{tab:ann_cost}
\begin{tabular}{lcc}
\toprule
\textbf{Component} & \textbf{Per image} & \textbf{Per dataset} \\
\midrule
Embedding computation & $0.050$ s & $\sim140{,}800$ s (39.1 h) \\
ANN retrieval (CPU)   & $0.003$ s & $\sim3{,}840$ s (1.07 h) \\
\bottomrule
\end{tabular}
\end{table}

While memory usage grows linearly with the number of stored embeddings and may
become challenging at very large scales, the runtime overhead of ANN retrieval
remains negligible compared to embedding computation, contributing only a minor
fraction of the total query time.

\subsection{Runtime and memory comparison}
\label{app:time_comparison}
Here we report the computational cost,
query-time overhead, and memory requirements for OOD scoring using a ResNet101
backbone for classic OOD methods. For the embedding density we use the DINO-V2 model. Unless stated otherwise, all dataset-
level measurements in Table~\ref{tab:time_memory_comparison} are reported for CPU
execution.

\begin{table*}[t]
\centering
\footnotesize
\captionsetup{width=\textwidth}
\setlength{\tabcolsep}{4pt}
\renewcommand{\arraystretch}{1.15}

\caption{Runtime, query-time overhead, and memory requirements for OOD methods
using a ResNet101 backbone on CPU. Preprocessing time is reported per image
(excluding the total preprocessing time) and corresponds to offline evaluation
over the full dataset. These timings reflect the full computation required to
obtain the OOD score for a single image and therefore apply uniformly whenever
the score is evaluated. (Mahalanobis: 15k images; others: $\sim$1.28M images).}

\begin{tabular}{
p{2.8cm}  
p{2.2cm}  
p{2.0cm}  
p{2.8cm}  
p{2.0cm}  
p{2.6cm}  
}
\toprule
\textbf{Method} &
\textbf{\# Preproc Samples} &
\textbf{Preproc / img (s)} &
\textbf{Total preproc time (h)} &
\textbf{Query overhead (s)} &
\textbf{Extra memory} \\
\midrule

Energy
& $\sim1.28$M
& $\sim0.15$
& $\sim53.3$
& $\sim0$
& $\sim1$ GB \\

Mahalanobis
& 15k
& $\sim0.05$
& $\sim0.2$
& $\sim0$
& $\sim400$ MB \\

ODIN
& $\sim1.28$M
& $\sim0.2$
& $\sim71.4$
& $\sim0.01$
& $\sim1$ GB \\

\midrule
Embedding Density (Ours)
& $\sim1.28$M
& $\sim0.055$
& $\sim39.1$
& $\sim0.003$
& $\sim8$ GB (dim = 768) \\

\bottomrule
\end{tabular}

\label{tab:time_memory_comparison}
\end{table*}

For Mahalanobis, preprocessing is performed on a reduced subset of 15k in-distribution images to estimate class-conditional feature statistics, namely per-class means and a shared covariance matrix. Once computed, OOD scoring for a queried image requires only a forward pass through the backbone followed by a closed-form distance computation, resulting in negligible query-time overhead. Since only a small set of statistics is stored, both the inference-time cost and persistent memory usage are independent of the dataset size.

Energy and ODIN do not require dataset-level statistics beyond those of the trained classifier. Preprocessing corresponds to offline evaluation over the full dataset for reporting purposes, while inference consists of a forward pass through the backbone followed by lightweight score computation. ODIN additionally performs a backward pass with respect to the input to compute the perturbed score, leading to a modest increase in per-image computation. In both cases, memory usage is dominated by the backbone parameters and transient runtime buffers, with no persistent per-image storage.

The embedding density method follows the same backbone-based embedding extraction procedure during preprocessing and inference, resulting in a comparable per-image computational cost for feature extraction. At query time, an additional overhead is incurred by approximate nearest-neighbor search over the stored embedding table, followed by a simple thresholding operation; this overhead remains small relative to the backbone forward pass. Unlike parametric methods, embedding density requires storing dataset-level representations: precomputed 768-dimensional embeddings extracted using the DINO-V2 backbone, as well as associated density statistics. In addition to storing the base embeddings, the method stores auxiliary density statistics of comparable size, resulting in a total memory footprint of 8GB, consistent with Table~\ref{tab:time_memory_comparison}.
\end{document}